# Wazobia Eval: A Benchmark for Nigerian Pidgin Emotion Understanding, Sarcasm Detection, and Cultural Reasoning

Stephanie Okoye
*Wazobia Labs*
*Lagos, Nigeria*
wazobialabs@gmail.com

**Abstract**

Nigerian Pidgin is one of Africa's most widely spoken languages, yet remains severely underrepresented in language model evaluation. Existing benchmarks primarily focus on translation, transcription, or generic sentiment analysis, leaving critical aspects of culturally grounded language understanding unmeasured. We introduce Wazobia Eval, a benchmark for evaluating Nigerian Pidgin emotion understanding, sarcasm detection, and cultural reasoning. The benchmark is built on a manually annotated dataset containing over 550 examples and a 16-category emotion taxonomy designed to capture culturally specific emotional registers that are not represented in conventional sentiment frameworks. Wazobia Eval provides standardized evaluation protocols and benchmark tasks for assessing model performance on nuanced Nigerian language understanding. We present the benchmark design, annotation methodology, taxonomy development process, and preliminary pilot evaluation results. Our goal is to provide foundational evaluation infrastructure for Nigerian language AI and establish a reproducible benchmark for future research. The dataset is publicly available at https://huggingface.co/WAZOBIALABS.

## 1. Introduction

Large language models have demonstrated impressive performance across a wide range of languages and tasks. However, the majority of existing evaluation benchmarks remain concentrated on high-resource languages, particularly English, leaving many African languages and language varieties insufficiently represented in modern AI evaluation frameworks. Nigerian Pidgin, one of the most widely spoken languages in West Africa, is a notable example of this gap.

While Nigerian Pidgin serves as a primary or secondary language for tens of millions of speakers across Nigeria and the broader West African region, language technologies for Nigerian Pidgin remain underdeveloped. Existing research has primarily focused on speech recognition, machine translation, and basic sentiment analysis. Comparatively little attention has been given to evaluating whether language models can understand the emotional, cultural, and pragmatic dimensions of Nigerian Pidgin communication.

This limitation is particularly significant because emotional expression in Nigerian Pidgin often relies on culturally specific meanings that do not map cleanly onto conventional sentiment categories such as positive, negative, and neutral. Expressions such as "I no fit shout," "abeg leave matter," or "God go run am" frequently convey emotional states that depend heavily on cultural context, social norms, and shared lived experiences. As a result, models that perform adequately on standard sentiment benchmarks may still fail to capture the intended meaning of Nigerian Pidgin utterances.

The challenge extends beyond emotion recognition. Nigerian Pidgin communication frequently incorporates sarcasm, indirect speech, social signaling, market-oriented interactions, religious expressions, and culturally grounded reasoning patterns. These phenomena are rarely represented in existing multilingual benchmarks, creating a substantial blind spot in the evaluation of language models intended for African users.

To address this gap, we introduce Wazobia Eval, a benchmark for evaluating Nigerian Pidgin emotion understanding, sarcasm detection, and cultural reasoning. The benchmark is built upon a manually annotated dataset containing more than 550 examples and a novel 16-category Nigerian emotion taxonomy designed to capture emotional registers that are common in Nigerian communication but absent from conventional sentiment frameworks.

Wazobia Eval is motivated by the belief that progress in Nigerian language AI requires not only datasets but also robust evaluation infrastructure. Datasets alone cannot determine whether a model truly understands a language. Standardized benchmarks are necessary to measure capability, compare systems, identify weaknesses, and track progress over time. Inspired by influential evaluation frameworks such as GLUE, SuperGLUE, and MMLU, Wazobia Eval seeks to provide a reproducible and extensible benchmark for Nigerian Pidgin language understanding.

The contributions of this work are threefold. First, we introduce a culturally grounded emotion taxonomy for Nigerian Pidgin that extends beyond traditional sentiment analysis. Second, we present a benchmark dataset and evaluation framework covering emotion classification, sarcasm detection, and cultural reasoning. Third, we establish a standardized evaluation protocol that enables systematic comparison of language models on Nigerian Pidgin understanding tasks.

Our broader goal is to contribute foundational evaluation infrastructure for Nigerian language AI and to encourage the development of language technologies that better reflect the linguistic and cultural realities of African users.

## 2. Related Work

### 2.1 Benchmarking Language Understanding

Benchmarking has played a central role in measuring progress in natural language processing. Resources such as GLUE and SuperGLUE established standardized evaluation frameworks for comparing language models across multiple tasks, enabling consistent measurement of language understanding capabilities.

More recently, large-scale benchmarks such as MMLU have expanded evaluation beyond traditional NLP tasks toward broader reasoning and knowledge assessment. These benchmarks have significantly influenced model development by identifying capability gaps and providing common evaluation standards.

However, most widely used benchmarks primarily focus on English and other high-resource languages, limiting their ability to assess performance in underrepresented linguistic contexts.

### 2.2 Emotion Understanding and Sentiment Analysis

Emotion classification and sentiment analysis have been extensively studied within NLP. Shared tasks such as SemEval have contributed benchmark datasets for emotion detection, sentiment classification, and affective computing.

While these resources have advanced emotion understanding research, many are based on linguistic and cultural assumptions that do not fully transfer to African language contexts. Emotion categories commonly used in existing benchmarks may fail to capture culturally specific communicative patterns present in Nigerian discourse.

Wazobia Eval addresses this limitation through a culturally grounded emotion taxonomy developed specifically for Nigerian Pidgin.

### 2.3 African Language NLP

Recent years have seen growing interest in African language technologies. Community-driven initiatives such as Masakhane have contributed significantly to African NLP through dataset creation, machine translation resources, and collaborative research. Similarly, benchmark efforts such as AfriSenti have expanded sentiment analysis resources for African languages and highlighted the importance of culturally relevant evaluation.

These initiatives demonstrate increasing recognition of the need for language resources that reflect African linguistic realities. Wazobia Eval builds upon this broader movement by focusing specifically on Nigerian Pidgin emotion understanding, sarcasm detection, and cultural reasoning.

### 2.4 Evaluation Infrastructure for Underrepresented Languages

Despite growing investment in African language AI, evaluation infrastructure remains limited relative to that available for high-resource languages. In many cases, model performance is measured using translated benchmarks or tasks that do not adequately reflect local communicative practices.

As a result, models may appear capable on conventional evaluations while still failing to understand culturally grounded expressions, pragmatic meaning, or context-dependent language use. Wazobia Eval contributes toward addressing this gap by providing an open benchmark specifically designed to evaluate culturally grounded Nigerian Pidgin understanding. The benchmark emphasizes emotional interpretation, sarcasm recognition, and contextual reasoning rather than translation alone.

## 3. Nigerian Pidgin Emotion Taxonomy

A central challenge in evaluating Nigerian Pidgin language understanding is that conventional sentiment analysis frameworks fail to capture many emotionally meaningful distinctions present in everyday communication. Most existing datasets rely on coarse sentiment categories such as positive, negative, and neutral. While these categories may be sufficient for broad sentiment estimation, they are often inadequate for representing the emotional and cultural nuances expressed in Nigerian Pidgin.

During dataset construction, we observed that many utterances could not be accurately described using traditional sentiment labels. Expressions frequently conveyed culturally specific emotional states shaped by shared social experiences, economic realities, religious practices, interpersonal dynamics, and communication norms. As a result, we developed a Nigerian Pidgin emotion taxonomy consisting of sixteen categories designed to better reflect how emotions are expressed and interpreted within Nigerian contexts.

Unlike conventional sentiment frameworks, the proposed taxonomy emphasizes communicative intent, social meaning, and culturally grounded emotional states rather than simple polarity.

| Category | Definition |
|---|---|
| **Joy** | Happiness, delight, positive emotional excitement |
| **Celebration** | Achievement, success, congratulations, festive events |
| **Anger** | Frustration, irritation, resentment, outrage |
| **Hustle Fatigue** | Exhaustion / burnout from prolonged economic struggle |
| **Hustle Energy** | Determination, ambition, persistence despite challenges |
| **Market Energy** | Transactional confidence, negotiation-oriented assertiveness |
| **Suspicion** | Doubt, mistrust, skepticism about another's motives |
| **Shock** | Surprise, disbelief, reaction to unexpected events |
| **Craving** | Strong desire, longing, anticipation |
| **Forming** | Deliberate performance of indifference or composure |
| **Betrayal** | Hurt, disappointment, violation of trust |
| **Pride** | Self-respect, dignity, accomplishment, identity affirmation |
| **Contempt** | Dismissive or scornful attitude toward a person/idea |
| **Prayer Gratitude** | Religious appreciation, blessing, divine acknowledgment |
| **Sarcasm** | Intended meaning differs from literal; irony / mock praise |
| **Neutral** | No strong expression of any preceding category |

*Figure 1. Wazobia Emotion Taxonomy — 16 Categories*

### 3.1 Taxonomy Development

The taxonomy was developed through iterative annotation and review of Nigerian Pidgin text examples collected across multiple domains. Categories emerged from repeated annotation challenges where existing emotion labels failed to adequately represent the intended meaning of an utterance.

The objective was not to create an exhaustive theory of emotion, but rather to establish a practical evaluation framework capable of measuring whether language models understand emotionally significant distinctions commonly encountered in Nigerian Pidgin communication.

### 3.2 Emotion Categories

**Joy.** Expressions of happiness, delight, satisfaction, or positive emotional excitement.

**Celebration.** Emotion associated with achievement, success, congratulations, public recognition, or festive events.

**Anger.** Direct expressions of frustration, irritation, resentment, outrage, or confrontation.

**Hustle Fatigue.** A culturally grounded emotional state characterized by exhaustion, burnout, discouragement, or emotional depletion resulting from prolonged economic or social struggle.

**Hustle Energy.** Expressions of determination, ambition, motivation, persistence, and commitment to continued effort despite challenges.

**Market Energy.** Transactional confidence, negotiation-oriented assertiveness, opportunism, or competitive social interaction commonly associated with commercial exchanges.

**Suspicion.** Doubt, mistrust, skepticism, caution, or concern regarding another person's motives or claims.

**Shock.** Surprise, disbelief, astonishment, or emotional reaction to unexpected events or information.

**Craving.** Strong desire, longing, anticipation, or yearning for an object, experience, or outcome.

**Forming.** Deliberate performance of indifference, composure, status, or emotional restraint. The speaker intentionally presents themselves as unconcerned or unaffected.

**Betrayal.** Feelings of disappointment, hurt, abandonment, or violation of trust by another person or institution.

**Pride.** Expressions of self-respect, confidence, dignity, accomplishment, or affirmation of identity.

**Contempt.** Dismissive, belittling, or scornful attitudes toward a person, idea, or situation.

**Prayer Gratitude.** Emotion expressed through religious appreciation, thanksgiving, blessing, testimony, or acknowledgment of divine intervention.

**Sarcasm.** Statements in which the intended meaning differs from the literal interpretation, often involving irony, mock praise, ridicule, or indirect criticism.

**Neutral.** Utterances that do not strongly express any of the preceding emotional categories.

### 3.3 Cultural Relevance

Several categories within the taxonomy do not have direct equivalents in standard emotion classification benchmarks. In particular, hustle_fatigue, hustle_energy, market_energy, forming, and prayer_gratitude emerged from recurring patterns in Nigerian communication and reflect culturally meaningful emotional registers that are underrepresented in existing NLP resources.

These categories were retained because they capture distinctions that are important for real-world language understanding and are frequently encountered in conversational Nigerian Pidgin. Their inclusion enables Wazobia Eval to assess dimensions of language understanding that would otherwise remain invisible under traditional sentiment classification frameworks.

### 3.4 Benchmark Implications

The taxonomy forms the foundation of the emotion classification task within Wazobia Eval. By requiring models to distinguish between sixteen emotion categories rather than broad sentiment labels, the benchmark provides a more demanding and culturally grounded evaluation of Nigerian Pidgin understanding.

The taxonomy also serves as a framework for future expansion into additional Nigerian languages and dialects, supporting the long-term goal of developing evaluation infrastructure for African language AI.

## 4. Dataset Construction

### 4.1 Data Collection

The Wazobia Eval dataset was constructed to address the lack of culturally grounded evaluation resources for Nigerian Pidgin. Data were collected from naturally occurring Nigerian Pidgin expressions and conversations representing everyday communication contexts.

Examples were designed to capture linguistic variation across multiple domains including social interaction, greetings, commerce, health communication, emotional expression, and culturally specific conversational exchanges.

The dataset emphasizes authentic Nigerian communicative patterns rather than translated English sentences. Particular attention was paid to preserving pragmatic meaning, local expressions, discourse markers, and culturally embedded language use.

### 4.2 Annotation Framework

Each example was manually annotated using a custom Nigerian emotion taxonomy consisting of sixteen emotion categories developed specifically for Nigerian communicative contexts.

The taxonomy includes both conventional emotional states and culturally specific emotional registers frequently observed in Nigerian discourse but insufficiently represented by standard sentiment analysis frameworks. Examples include categories such as hustle_fatigue, hustle_energy, market_energy, forming, and prayer_gratitude.

The annotation process prioritized contextual meaning rather than lexical cues alone.

### 4.3 Sarcasm Dataset Construction

To evaluate pragmatic language understanding, a dedicated sarcasm subset was created. Sarcastic expressions were paired with semantically similar sincere counterparts, allowing controlled evaluation of a model's ability to distinguish literal meaning from intended meaning.

This approach reduces reliance on surface-level lexical patterns and encourages evaluation of contextual reasoning capabilities.

### 4.4 Context Dependency and Ambiguity

A notable characteristic of Nigerian Pidgin communication is that emotional meaning is often highly dependent on context rather than lexical content alone.

During annotation, several expressions were observed to convey different emotional meanings despite having identical surface forms. For example, the expression "You don try well well" may represent celebration, pride, sarcasm, or contempt depending on speaker intent, conversational context, and interpersonal dynamics.

Similarly, many Nigerian Pidgin utterances rely on shared cultural knowledge, tone, social relationships, and situational context for interpretation. As a result, annotation decisions prioritized intended meaning rather than literal wording.

**Example 1. Context-Dependent Emotional Meaning**

Utterance: *"You don try well well."*

Context A: A student becomes the first person in their family to graduate. **Label: celebration**
Context B: A mechanic damages a customer's vehicle. **Label: sarcasm**
Context C: A politician fails to fulfill campaign promises. **Label: contempt**

This example illustrates how identical Nigerian Pidgin utterances may express different emotional meanings depending on context and speaker intent.

This observation highlights an important challenge for language model evaluation. Models must move beyond keyword matching and demonstrate contextual understanding in order to correctly interpret emotionally ambiguous expressions.

The presence of context-dependent emotional meaning motivates future benchmark tracks focused on contextual emotion disambiguation and pragmatic reasoning.

### 4.5 Dataset Statistics

The current release contains more than 550 annotated examples. The dataset includes 16 emotion categories, 28 sarcasm pairs, health-domain examples, social interaction examples, commerce-related examples, and gender representation across speakers.

| Category | Count |
|---|---|
| Total Annotated Examples | 550+ |
| Emotion Categories | 16 |
| Sarcasm Pairs | 28 |
| Benchmark-Ready Examples | 253 |
| Pilot Benchmark Examples | 16 |
| License | CC BY 4.0 |
| HuggingFace | https://huggingface.co/WAZOBIALABS |

*Table 1. Dataset Summary*

### 4.6 Intended Use

The dataset is intended for benchmark development, emotion classification, sarcasm detection, cultural reasoning evaluation, Nigerian Pidgin NLP research, and African language AI evaluation. The dataset is not intended to represent all varieties of Nigerian Pidgin or all Nigerian cultural perspectives. Instead, it serves as an initial benchmark resource designed to support future research and evaluation efforts. The dataset is publicly available under the Creative Commons Attribution 4.0 license at https://huggingface.co/WAZOBIALABS.

## 5. Wazobia Eval Benchmark Design

### 5.1 Benchmark Overview

Wazobia Eval is designed to evaluate whether language models can understand Nigerian Pidgin beyond literal translation and surface-level sentiment classification. The benchmark focuses on emotional understanding, pragmatic reasoning, and culturally grounded interpretation.

The benchmark consists of three primary evaluation tracks: (1) Emotion Classification, (2) Sarcasm Detection, and (3) Cultural Reasoning. Together, these tasks assess both linguistic competence and contextual understanding.

**WAZOBIA EVAL**

Track A: Emotion Classification | Track B: Sarcasm Detection | Track C: Cultural Reasoning

Standardized Prompt Protocol

GPT · Claude · Gemini · Open-Source

Accuracy · Precision · Recall · F1 · Macro-F1

*Figure 2. Wazobia Eval Benchmark Architecture*

### 5.2 Track A: Emotion Classification

**Objective.** Given a Nigerian Pidgin utterance, predict the most appropriate emotion label from the Wazobia emotion taxonomy.

**Input.** A single Nigerian Pidgin utterance, e.g., *"I no fit shout again."*

**Output.** Exactly one emotion label from the sixteen-category taxonomy, e.g., *hustle_fatigue*.

**Evaluation Metrics.** The primary metric is Macro-F1 Score. Additional metrics include Accuracy, Precision, Recall, and Per-Class F1. Macro-F1 is emphasized because it gives equal importance to all emotion categories and reduces bias toward majority classes.

### 5.3 Track B: Sarcasm Detection

**Objective.** Determine whether an utterance is sarcastic or non-sarcastic.

**Input.** A Nigerian Pidgin utterance, e.g., *"You try well well."*

**Output.** One of two labels: Sarcastic or Not Sarcastic.

**Evaluation Metrics.** Accuracy, Precision, Recall, and F1 Score. This task evaluates a model's ability to distinguish intended meaning from literal meaning.

### 5.4 Track C: Cultural Reasoning

**Objective.** Evaluate whether a model can explain culturally grounded meanings and intentions embedded within Nigerian Pidgin expressions.

**Input.** A Nigerian Pidgin utterance, e.g., *"You dey form."*

**Output.** A short explanation describing the intended meaning in context.

**Evaluation.** Outputs are assessed using qualitative review and human evaluation. The focus is on contextual correctness rather than lexical similarity.

### 5.5 Evaluation Protocol

All evaluated models receive identical instructions and evaluation prompts. Models are not provided access to benchmark labels beyond the permitted output space. To ensure comparability, all models are evaluated using standardized prompts and identical benchmark examples. Each model is assessed independently and evaluated using the same scoring methodology.

### 5.6 Benchmark Philosophy

Wazobia Eval is designed to measure understanding rather than translation. A model may successfully translate a Nigerian Pidgin sentence into English while still failing to identify its emotional or cultural meaning. The benchmark therefore prioritizes cultural competence, pragmatic understanding, and emotionally grounded interpretation. This philosophy reflects the broader goal of developing evaluation infrastructure that more accurately measures language understanding for Nigerian and African language communities.

## 6. Evaluation Protocol

### 6.1 Evaluated Models

Wazobia Eval is designed as a model-agnostic benchmark that can be applied to a wide range of language models. The benchmark framework supports evaluation of both proprietary and open-source systems.

For the pilot benchmark reported in this paper, evaluation was conducted using GPT-5.5 under standardized prompting conditions. Future benchmark releases will include comparative evaluations across multiple frontier and open-source models, including systems developed by OpenAI, Anthropic, Google, Meta, and other organizations.

### 6.2 Prompt Design

All models were evaluated using a standardized prompt protocol to ensure comparability across systems. For the emotion classification task, models received a Nigerian Pidgin utterance together with the complete Wazobia emotion taxonomy and were instructed to select exactly one label. Models were explicitly instructed not to provide explanations, reasoning traces, or additional text beyond the selected label. This constraint reduced output variability and simplified automated scoring.

The prompt protocol was designed to minimize prompt-induced performance differences and ensure that benchmark scores reflected model understanding rather than prompt engineering advantages. All evaluated models received identical instructions and identical benchmark examples.

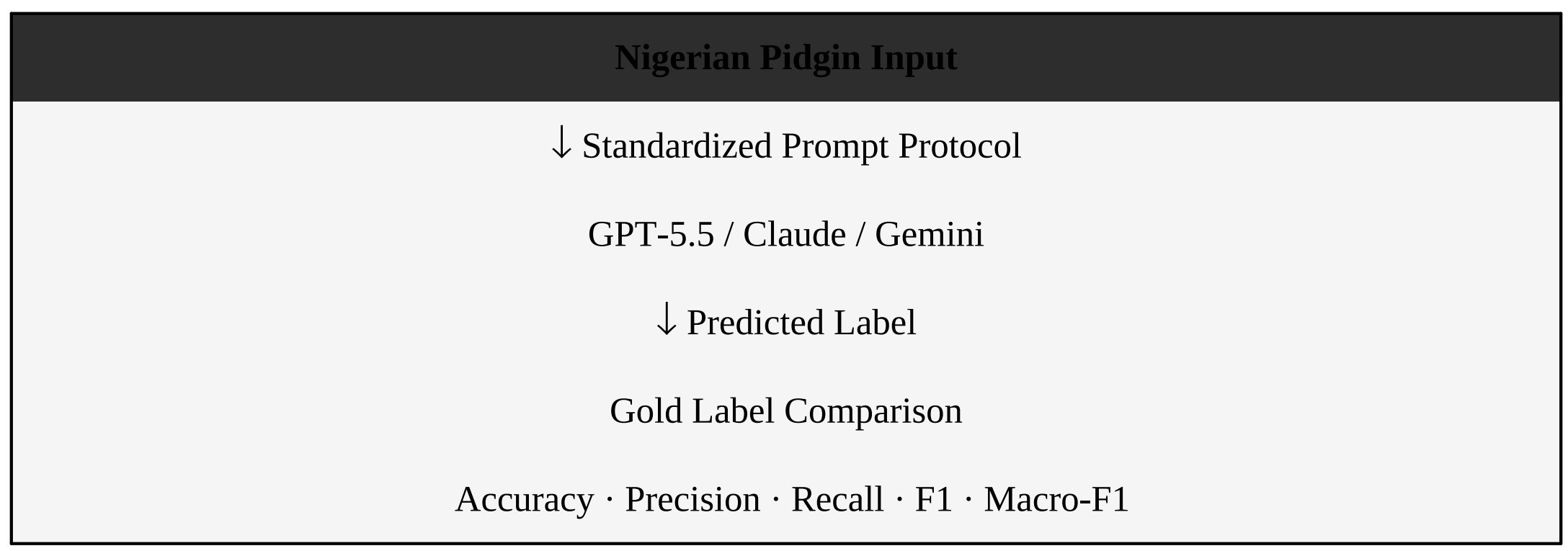

*Figure 3. Wazobia Evaluation Workflow*

### 6.3 Evaluation Dataset

For the initial release of Wazobia Eval, benchmark examples were drawn from a manually annotated Nigerian Pidgin dataset containing more than 550 examples spanning emotion classification, sarcasm detection, and culturally grounded language understanding tasks.

A dedicated evaluation subset was constructed to support reproducible benchmark assessment. The evaluation set contains 253 benchmark-ready examples distributed across the sixteen emotion categories defined in the Wazobia emotion taxonomy.

To validate benchmark procedures prior to large-scale evaluation, a pilot benchmark was conducted using a balanced subset containing representative examples from each emotion category. This pilot benchmark served as an initial test of prompt design, scoring methodology, evaluation workflows, and result recording procedures.

The evaluation dataset is intended to evolve over time through additional annotation, quality review, and benchmark expansion. Future releases of Wazobia Eval will introduce larger benchmark splits, contextual reasoning tasks, and ambiguity-focused evaluation tracks.

### 6.4 Evaluation Metrics

Wazobia Eval uses standard classification metrics to enable reproducible comparison across language models.

**Accuracy** measures the proportion of correctly classified examples relative to the total number of evaluation examples.

**Precision** measures the proportion of predicted labels that are correct. This metric is particularly useful when evaluating categories that may be confused with semantically related emotional states.

**Recall** measures the proportion of true examples that are successfully identified by a model.

**F1 Score** combines precision and recall into a single balanced measure of classification performance.

**Macro-F1** is the primary evaluation metric used in Wazobia Eval. Because the benchmark contains multiple emotion categories with varying frequencies, Macro-F1 assigns equal weight to each category and prevents dominant classes from disproportionately influencing overall performance.

### 6.5 Scoring Procedure

All benchmark examples are evaluated using standardized prompts and identical evaluation conditions. For the emotion classification task, a model receives a Nigerian Pidgin utterance and is required to predict exactly one emotion label from the Wazobia emotion taxonomy. Predicted labels are compared against gold-standard human annotations. Correct predictions receive a score of one, while incorrect predictions receive a score of zero. Aggregate benchmark scores are then computed using Accuracy, Precision, Recall, F1 Score, and Macro-F1.

For sarcasm detection, model outputs are evaluated against manually annotated sarcasm labels. For cultural reasoning tasks, outputs are reviewed using qualitative assessment criteria focused on contextual correctness, cultural interpretation, and pragmatic understanding.

## 7. Pilot Benchmark Results

### 7.1 Pilot Evaluation

To validate the benchmark workflow, a pilot evaluation was conducted using a balanced subset containing sixteen examples representing all emotion categories within the Wazobia emotion taxonomy.

The pilot benchmark was evaluated using GPT-5.5 under the standardized prompt protocol described in Section 6. The primary objective of the pilot evaluation was not to establish definitive benchmark scores, but rather to validate benchmark procedures, prompt design, scoring workflows, and result recording processes.

### 7.2 Preliminary Results

GPT-5.5 correctly classified 7 out of 16 benchmark examples, corresponding to an accuracy of 43.75%.

While the pilot benchmark is too small to support strong statistical conclusions, the results provide initial evidence that culturally grounded Nigerian Pidgin emotion understanding remains challenging for contemporary language models. Several prediction errors involved categories that rely heavily on cultural context, pragmatic interpretation, or distinctions between emotionally similar labels.

### 7.3 Observed Challenges

Analysis of model outputs revealed several recurring challenges:

- Difficulty distinguishing between culturally specific emotion categories.
- Confusion between sincere praise and sarcastic expressions.
- Misclassification of context-dependent utterances.
- Reduced performance on expressions requiring cultural knowledge beyond lexical meaning.

These observations support the motivation behind Wazobia Eval and suggest that conventional language model evaluation benchmarks may not adequately capture culturally grounded understanding in Nigerian Pidgin.

### 7.4 Limitations of the Pilot Evaluation

The pilot benchmark was intentionally small and designed primarily for workflow validation. Additionally, because evaluation was conducted within conversational interfaces, contextual carryover effects may have influenced some model predictions. Future benchmark evaluations will use isolated evaluation sessions and larger benchmark subsets to improve reliability and reproducibility.

The results presented here should therefore be interpreted as preliminary findings rather than definitive measurements of model capability.

### 7.5 Error Analysis

Analysis of model predictions revealed several recurring error patterns.

First, the model frequently confused emotionally related categories such as joy, celebration, and pride. While the model often recognized positive sentiment, it struggled to distinguish between different forms of positive emotional expression.

Second, the model exhibited difficulty with culturally grounded categories including forming, contempt, and sarcasm. These categories often depend on contextual information, speaker intent, and pragmatic interpretation rather than lexical content alone.

Third, several benchmark examples demonstrated the importance of contextual reasoning. Expressions such as "You don try well well" may convey celebration, sincere praise, sarcasm, or contempt depending on conversational context. In the absence of explicit contextual information, these examples remain challenging even for advanced language models.

Finally, the model occasionally confused culturally specific categories with more general emotional concepts. For example, examples labeled as market_energy were sometimes interpreted as craving, suggesting that additional label definitions and benchmark guidance may improve evaluation consistency.

These findings support the central motivation of Wazobia Eval: measuring culturally grounded language understanding rather than translation or surface-level sentiment recognition.

## 8. Discussion

The results of the pilot benchmark highlight the broader challenge of evaluating language understanding in underrepresented languages and language varieties. While recent large language models have demonstrated impressive performance on widely used benchmarks, the preliminary findings from Wazobia Eval suggest that culturally grounded Nigerian Pidgin understanding remains a difficult task.

A key observation from this work is that many failures are not caused by a lack of vocabulary knowledge. Instead, errors often arise from difficulties in interpreting cultural context, pragmatic intent, sarcasm, social signaling, and emotionally ambiguous expressions. This finding supports the argument that language understanding extends beyond translation and lexical recognition.

The Wazobia emotion taxonomy also illustrates limitations in conventional sentiment analysis frameworks. Categories such as hustle_fatigue, hustle_energy, market_energy, forming, and prayer_gratitude represent communicative patterns that are common in Nigerian discourse but rarely appear in existing benchmark resources. As a result, models may appear capable when evaluated on standard sentiment tasks while still failing to capture important culturally grounded distinctions.

Another important observation is the prevalence of context-dependent emotional meaning. During annotation, several Nigerian Pidgin expressions were found to convey different emotional interpretations depending on speaker intent and conversational context. These examples expose weaknesses in evaluation approaches that rely primarily on lexical matching and suggest the need for future benchmarks focused on contextual reasoning and pragmatic understanding.

### 8.1 Future Benchmark Expansion: Contextual Emotion Disambiguation

One of the most significant findings during dataset construction was the prevalence of context-dependent emotional meaning. Several Nigerian Pidgin expressions were observed to convey multiple emotional interpretations despite having identical lexical forms.

This observation motivates a future benchmark track focused on Contextual Emotion Disambiguation. Unlike traditional emotion classification tasks, this track would require models to infer emotional meaning using both the utterance and contextual information. The objective is to evaluate whether language models can move beyond lexical recognition and demonstrate deeper pragmatic understanding of Nigerian Pidgin communication.

The broader goal of Wazobia Eval is therefore not simply to introduce another dataset, but to establish a foundation for culturally grounded language evaluation. Future benchmark expansions may include additional Nigerian languages, multilingual evaluation tracks, contextual disambiguation tasks, safety assessments, and human-model comparison studies.

## 9. Limitations

Several limitations should be considered when interpreting the findings presented in this work.

First, the current version of Wazobia Eval represents an initial benchmark release and is limited in scale relative to established language evaluation benchmarks. Although the dataset contains more than 550 annotated examples, additional data collection and annotation efforts will be necessary to support larger and more comprehensive evaluations.

Second, the benchmark currently focuses primarily on Nigerian Pidgin and therefore does not capture the full linguistic diversity of Nigeria or the African continent. Future benchmark expansions may incorporate additional Nigerian languages and multilingual evaluation settings.

Third, while the dataset was manually annotated using a culturally grounded emotion taxonomy, formal inter-annotator agreement analysis has not yet been completed for the full dataset. Future work will include independent annotation studies and agreement measurements to further validate label consistency and reproducibility.

Fourth, many Nigerian Pidgin expressions derive meaning from context, tone, speaker relationships, and situational factors that are difficult to fully capture in text-only datasets. As a result, some emotionally ambiguous examples may admit multiple plausible interpretations depending on context.

Fifth, the pilot benchmark results reported in this paper are preliminary and were intended primarily to validate benchmark workflows rather than establish definitive model rankings. Larger evaluation sets and controlled benchmarking procedures will be required to support stronger conclusions regarding model performance.

Despite these limitations, Wazobia Eval provides an initial foundation for culturally grounded Nigerian language evaluation and establishes a framework that can be expanded through future research and community contributions.

## 10. Conclusion

This paper introduced Wazobia Eval, a benchmark for evaluating emotion understanding, sarcasm detection, and cultural reasoning in Nigerian Pidgin.

To support this benchmark, we developed a culturally grounded Nigerian emotion taxonomy containing sixteen categories that extend beyond conventional sentiment analysis frameworks. The benchmark was constructed using a manually annotated dataset of more than 550 Nigerian Pidgin examples and includes dedicated evaluation tracks designed to assess emotional understanding, pragmatic interpretation, and culturally informed reasoning.

Preliminary pilot evaluations suggest that culturally grounded Nigerian Pidgin understanding remains challenging for contemporary language models, particularly when interpretation depends on context, sarcasm, or culturally specific emotional concepts. These findings highlight the need for evaluation resources that move beyond translation and lexical recognition toward deeper measures of language understanding.

More broadly, this work contributes to ongoing efforts to improve evaluation infrastructure for African language AI. As language technologies become increasingly important across education, healthcare, government, and commerce, reliable benchmarks will play a critical role in measuring progress and identifying capability gaps.

Wazobia Eval represents an initial step toward that goal. Future work will expand benchmark coverage, introduce contextual disambiguation tasks, incorporate additional African languages, establish formal inter-annotator agreement studies, and support broader community participation in benchmark development.

By providing open evaluation infrastructure for Nigerian Pidgin, Wazobia Eval aims to support the development of more inclusive, culturally aware, and representative language technologies. The dataset and benchmark are available at https://huggingface.co/WAZOBIALABS.